\PassOptionsToPackage{obeyspaces}{url}
\documentclass[sigconf,screen]{acmart}
\usepackage[subfig,acm]{definition}

\copyrightyear{2022} 
\acmYear{2022} 
\setcopyright{acmlicensed}
\acmConference[CIKM '22]{Proceedings of the 31st ACM International Conference on Information and Knowledge
Management}{October 17--21, 2022}{Atlanta, GA, USA}
\acmBooktitle{Proceedings of the 31st ACM International Conference on Information and Knowledge Management (CIKM '22), October 17--21, 2022, Atlanta, GA, USA}
\acmPrice{15.00}
\acmDOI{10.1145/3511808.3557577}
\acmISBN{978-1-4503-9236-5/22/10}

\newcommand{\themodel}{\textsf{CREME}\xspace}

\begin{document}

\title{Deep Contrastive Multiview Network Embedding}

\author[Mengqi Zhang, Yanqiao Zhu, Qiang Liu, Shu Wu, and Liang Wang]{Mengqi Zhang*, Yanqiao Zhu*, Qiang Liu, Shu Wu$^{\dagger}$, and Liang Wang}

\authornotetext{These authors made equal contribution to this work.}
\authornotetext{To whom correspondence should be addressed.}

\affiliation{%
	\institution{$^1$School of Artificial Intelligence, University of Chinese Academy of Sciences}
	\institution{$^2$Center for Research on Intelligent Perception and Computing, Institute of Automation, Chinese Academy of Sciences}
	\country{}
}

\email{{mengqi.zhang, yanqiao.zhu}@cripac.ia.ac.cn, {qiang.liu, shu.wu, wangliang}@nlpr.ia.ac.cn}

\def\authors{Mengqi Zhang, Yanqiao Zhu, Qiang Liu, Shu Wu, and Liang Wang}

\begin{abstract}
Multiview network embedding aims at projecting nodes in the network to low-dimensional vectors, while preserving their multiple relations and attribute information.
Contrastive learning approaches have shown promising performance in this task. However, they neglect the semantic consistency between fused and view representations and have difficulty in modeling complementary information between different views.
To deal with these deficiencies, this work presents a novel \underline{C}ontrastive lea\underline{R}ning fram\underline{E}work for \underline{M}ultiview network \underline{E}mbedding (\themodel).
In our work, different views can be obtained based on the various relations among nodes.
Then, we generate view embeddings via proper view encoders and utilize an attentive multiview aggregator to fuse these representations.
Particularly, we design two collaborative contrastive objectives, view fusion InfoMax and inter-view InfoMin, to train the model in a self-supervised manner.
The former objective distills information from embeddings generated from different views, while the latter captures complementary information among views to promote distinctive view embeddings.
We also show that the two objectives can be unified into one objective for model training.
Extensive experiments on three real-world datasets demonstrate that our proposed \themodel is able to consistently outperform state-of-the-art methods.
\end{abstract}

\ccsdesc[500]{Computing methodologies~Unsupervised learning}

\maketitle

\section{Introduction}
\label{intro}
Real-world networks often consist of various types of relations, which are known as multiview networks or multiplex networks.
Take academic networks as an example: Two nodes denoting papers are connected if they have common authors or they cite each other.
Multiview network embedding aims at projecting nodes in the network to low-dimensional vectors, while preserving their multiple relations and attribute information \cite{chu2019cross,zhang2018scalable,qu2017attention,Park:2020jj}.
Since it is usually labor-extensive to manually collect high-quality labels, obtaining informative embeddings without supervision for multiview networks has attracted a lot of attention in the community.

So far, a series of self-supervised methods have been proposed for multiview network embedding.
Some early approaches \cite{chu2019cross,zhang2018scalable,qu2017attention} mainly focus on the compression of multiple graph views but ignore node attributes.
To capture the attribute and structure information together, some others \cite{cen2019representation,ma2018multi} combine graph neural networks and relational reconstruction tasks for self-supervised learning.
However, most of these methods over-emphasize the network proximity, thus limiting the expressiveness of learned embeddings \cite{qiu2018network,ribeiro2017struc2vec,velivckovic2018deep}.
Inspired by visual representation learning \cite{hjelm2018learning}, recent studies attempt to introduce contrastive learning into multiview networks \cite{Park:2020jj,jing2021hdmi} and have achieved compelling performance.

However, we argue that these contrastive models still have two deficiencies.
Firstly, their contrastive strategies neglect the semantic consistency between views in the original network.
In the paradigm of multiview network embedding, the final node embedding is usually obtained by aggregating node embeddings from different views induced by relations.
Based on the hypothesis that a powerful representation is one that models view-invariant factors \cite{smith2005development,tian2020contrastive}, the fused embedding should capture sufficient semantic information shared among multiple relations.
In contrast, the existing models focus on contrasting node- and graph-level embeddings, while ignoring capturing view-invariant factors in relation-induced views. As a result, the fused representation suffers from limited expressiveness.
Secondly, these contrastive methods fail to further consider inter-view dependency, leading to suboptimal performance.
Consider that in multiview networks, node representations obtained from different views tend to be similar due to the shared node attributes.
To improve the discriminative ability of node embeddings, it is thus vital to capture the complementary information of different views \cite{Shi:2018vc}.

To deal with the two aforementioned challenges, we present a novel deep \underline{C}ontrastive lea\underline{R}ning fram\underline{E}work for \underline{M}ultiview network \underline{E}mbedding, \themodel for brevity.
The overall framework of \themodel is presented in \cref{fig:frame}.
Specifically, we first generate views according to various relations of multiview networks.
Then, we obtain each view representation via a view encoder based on graph attention networks (\S \ref{inner-view}).
Next, we combine all the relations and form a fusion view.
Accordingly, we introduce a multiview aggregator to integrate different view representations as the final node representations (\S \ref{multi-view}).
To enable self-supervised training, we propose a novel contrasting strategy (view fusion InfoMax) with a regularization term (inter-view InfoMin) (\S \ref{con_mv}).
The first objective \emph{maximizes the mutual information between the fused representation and view representations} to promote multiview fusion, while the second objective enforces \emph{information minimization among graph views}, which improves distinctiveness of view representations, so as to preserve complementary information among relation-induced views.
We further show that the two contrastive objectives can be collectively combined into one elegant, unified objective function.

The main contributions of this work are summarized as follows:
Firstly, we propose a novel contrastive framework \themodel for multiview network embedding, the core of which contains two collaborative contrastive objectives, view fusion InfoMax and inter-view InfoMin.
Secondly, we conduct extensive empirical studies on three real-world datasets. The results demonstrate the effectiveness of \themodel over state-of-the-art baselines.

\section{The Proposed Method}
\subsection{Preliminaries}

\noindent\textbf{Definition (Multiview networks).\enspace}
A multiview network is a graph $\mathcal{G}=(\mathcal{V},\mathcal{E}, \bm{X}, \phi)$ whose edges are associated with more than one types. In such a network, the mapping $\phi: \mathcal{E}\to \mathcal{R}, |\mathcal{R}| > 1$ associates an edge with an edge type; $\mathcal{V}$, $\mathcal{E} \in \mathcal{V} \times \mathcal{V} $, $\bm{X} \in \mathbb{R}^{|\mathcal{V}| \times F}$, and $\mathcal{R}$ represents the node set, the edge set, the node attribute matrix, and the set of edge types respectively. 

In this work, we consider the task of self-supervised multiview network embedding, where we aim to learn a  $d$-dimensional (\(d \ll F\)) vector \(\bm{z}_i\) representing for each node \(i\) without accessing to labels.

\subsection{The Overall Framework}

\begin{figure}[t]
	\centering
	\includegraphics[width=0.85\linewidth]{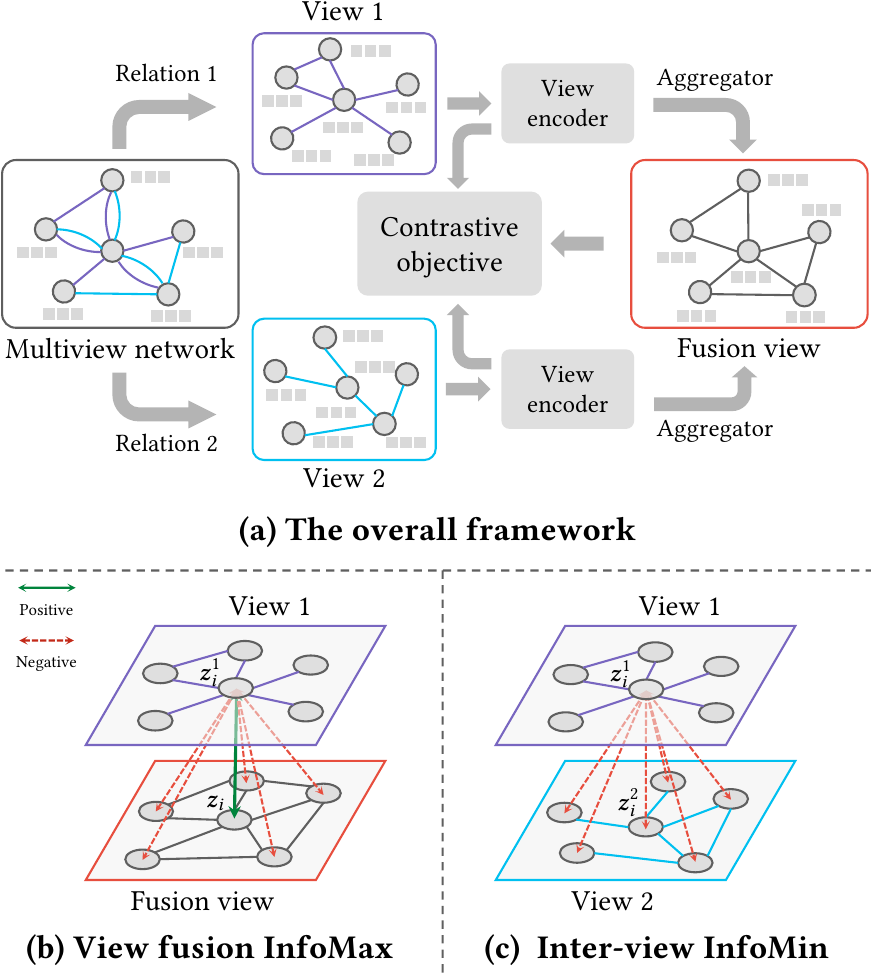}
	\caption{An overview of \themodel: Figure (a) presents the framework of our proposed model; Figures (b) and (c) illustrate the contrastive strategies between different graph views, where the green solid arrow indicates positive pairs and the orange dashed arrows indicate negative pairs.}
	\label{fig:frame}
\end{figure}

The overall framework of \themodel is illustrated in Figure \ref{fig:frame}.
There are three main components: (1) a view encoder that projects nodes in each relation-induced view into low-dimensional representations, (2) a multiview aggregator, which adaptively integrates view representations and obtains the final fused node embeddings for \(\mathcal{G}\), and (3) a unified contrastive objective to enable self-supervised learning of the view encoder and the multiview aggregator. 

Our \themodel framework follows the common multiview contrastive learning paradigm, which essentially seeks to maximize the agreement of representations among different views.
Different from traditional graph contrastive learning methods \cite{Velickovic:2019tu,you2020graph, hassani2020contrastive,zhu2021graph,Zhu:2020vf}, our graph views are naturally induced by different relations rather than generated by data augmentations.

After obtaining views according to relations, we utilize an encoding function $f_r: \bm{A}_r\times \bm{X} \to \bm{Z}^r \in \mathbb{R}^{N\times d}$ for view $\mathcal{G}_r$ to obtain relation view representations.
Thereafter, we employ a multiview aggregator $g: (\bm{Z}^1,...,\bm{Z}^{|\mathcal{R}|}) \to \bm{Z}\in \mathbb{R}^{N\times d}$ to obtain the fused representations for multiview network $\mathcal{G}$. 
Here, $\bm{z}^r_i$ in $\bm{Z}^r$ is the representation of node $i$ in view $\mathcal{G}_r$ and $\bm{z}_i$ in $\bm{Z}$ is the representation of node $i$ in graph $\mathcal{G}$, which can be regarded as a fusion view of the original network.
In this following subsections, we introduce the learning objective at first (\S \ref{con_mv}) and then proceed to the design of view encoders (\S \ref{inner-view}) and multiview aggregators (\S \ref{multi-view}) in detail.
 
\subsection{Contrastive Objectives}
\label{con_mv}

\subsubsection{View fusion InfoMax}
At first, we propose a novel contrasting strategy to train the model by maximizing the semantic consistency of view representation $\bm{Z}^r$ in each view $\mathcal{G}_r$ and the fusion representation \(\bm{Z}\) of $\mathcal{G}$.
Following mutual information estimation \cite{vandenOord:2018ut}, this can be achieved by maximizing the Mutual Information (MI) between $\bm{Z}^r$ and $\bm{Z}$. In this way, the resulting fusion view representation can \emph{selectively} distill information of each relation view. 

Specifically, for an anchor node \(i\), its view representation and the fused representation $(\bm{z}_i^r,\bm{z}_i)$ constitutes a positive pair.
Following prior studies \cite{Zhu:2020vf,Zhu:2021tu,zhu2021graph}, we set all other nodes in two graph views as negative pairs of $\bm{z}_i^r$.
We illustrate the view fusion InfoMax in Figure \ref{fig:frame}(b).
Formally, the objective of view fusion InfoMax is defined as
\begin{equation}
    \mathcal{L}_o(\bm{z}^r_i,\bm{z}_i)=\log\frac{e^{\theta(\bm{z}_i^r,\bm{z}_i)/\tau}}{e^{\theta(\bm{z}_i^r,\bm{z}_i)/\tau}+\sum_{j\neq i}e^{\theta(\bm{z}_i^r,\bm{z}_j)/\tau} + \sum_{j\neq i}e^{\theta(\bm{z}_i^r,\bm{z}_j^r)/\tau}},
    \label{loss1}
\end{equation}
where $\theta(u,v)=s(p(u),p(v))$ is a critic function, $s(\cdot,\cdot)$ is implemented using a simple cosine similarity, $p(\cdot)$ is a non-linear projection function to enhance the expression power of the critic function, and $\tau$ is a temperature parameter.
For simplicity, we denote the denominator in Eq.~(\ref{loss1}) as \(\rho(\bm{z}_i^r,\bm{z}_i)\) hereafter:
\begin{equation}
    \rho(\bm{z}_i^r,\bm{z}_i) = e^{\theta(\bm{z}_i^r,\bm{z}_i)/\tau}+\sum\nolimits_{j\neq i}e^{\theta(\bm{z}_i^r,\bm{z}_j)/\tau} + \sum\nolimits_{j\neq i}e^{\theta(\bm{z}_i^r,\bm{z}_j^r)/\tau}.
\end{equation}

\subsubsection{Inter-view InfoMin}
The previous objective only focuses on the relationship between each relation view and the final fusion view.
Considering that in our multiview network embedding setting, each node shares the same node attribute in different relation views, and thus their view embeddings tend to be similar during view encoding.
Therefore, we propose the second objective to further regularize information among relation views.
Our approach is to add a regularization term to minimize the MI of relation view representations, so as to enforce the model to learn discriminative view representations.

Instead of directly optimizing MI between $\bm{Z}^r$ and $\bm{Z}^k$ for any view pair \((r, k)_{r \neq k}\), we simply set $(z_i^r,z_i^k)_{r\neq k}$ and $(z_i^r,z_j^k)_{i\neq j}$ as additional negative samples in Eq.~(\ref{loss1}), as illustrated in Figure \ref{fig:frame}(c).
In this way, we elegantly combine the two contrastive objectives:
\begin{equation}
    \mathcal{L}(\bm{z}_i^r,\bm{z}_i)=\log\frac{e^{\theta(\bm{z}_i^r,\bm{z}_i)/\tau}}{\rho(\bm{z}_i^r,\bm{z}_i)+\sum_{j \in \mathcal{G}_k}{\mathds 1_{[k \neq r]}}e^{\theta(\bm{z}_i^r,\bm{z}_j^k)/\tau}}.
\end{equation}

\subsubsection{Learning objectives}
Finally, the overall objective is defined as an average of MI over all positive pairs, formally given by
\begin{equation}
    \mathcal{J} = \frac{1}{N\cdot |\mathcal{R}|}\sum\nolimits_{i=1}^N\sum\nolimits_{r=1}^{|\mathcal{R}|} \mathcal{L}(\bm{z}_i^r,\bm{z}_i).
\end{equation}

To summarize, the view fusion InfoMax objective enforces the multiview aggregator to adaptively distill information from each relation view.
The inter-view InfoMin regularization further constrains different relation view representations to be distinct to each other, which makes the model capture the complementary information of contained in each relation.

\subsection{View Encoders}
\label{inner-view}

For the input multiview network, we generate views, each according to one provided relation. In this way, we essentially convert the heterogeneous network into a series of homogeneous networks.
Then, for each relation-induced view, we capture the structural and attribute information of nodes through a view-specific graph attention neural network.
To be specific, we leverage the self-attention mechanism \cite{velivckovic2018graph} to compute the weight coefficients $\alpha_{ij}^r$ between node $i$ and its neighbor $j$ in $\mathcal{G}_r$: 
\begin{equation}
    \alpha_{ij}^{r} = \frac{\exp({\sigma(\bm{a}_r^\top [ \bm{M}_r\bm{x}_i \parallel  \bm{M}_r\bm{x}_j]}))}{\sum_{k\in \mathcal{N}_i^r}\exp({\sigma(\bm{a}_r^\top [ \bm{M}_r \bm{x}_i \parallel  \bm{M}_r\bm{x}_k]}))},
\end{equation}
where $\mathcal{N}_i^r$ is the set of neighbors of node $i$ in view $\mathcal{G}_r$, $\bm{a}_r \in \mathbb{R}^{2d}$ is a view-specific weight vector, $\bm{M}_r \in \mathbb{R}^{d\times F} $ is a transformation matrix projecting each node attribute into the corresponding semantic space, and $\sigma(\cdot) = \operatorname{ReLU}(\cdot)$ is the non-linear function. The view representation of node $i$ in $\mathcal{G}_r$ can then be calculated by aggregating their neighbor node embeddings:
\begin{equation}
\label{gat}
    \bm{z}_i^{r}= \concat\nolimits_{k = 1}^{K}\sigma\left(\sum\nolimits_{j\in \mathcal{N}_i^r}\alpha_{ij}^{r,k}\bm{M}_r\bm{x}_j\right).
\end{equation}
Here, we leverage a multihead attention mechanism of \(K\) heads \cite{velivckovic2018graph}, where each attention head has its own independent encoder parameters and \(\alpha_{ij}^{r,k}\) is the attention coefficient in the \(k\)-th head.

\subsection{Multiview Aggregators}
\label{multi-view}
After obtaining each relation view representations, we leverage a multiview aggregator to integrate semantic information from all relation-induced views for each node in $\mathcal{G}$.
To preserve important information during aggregation, we utilize an another attention network to aggregate the embeddings of different views for each node. The importance of each of view embedding $\bm{z}_i^r$ can be calculated by
\begin{equation}    
    w^r_i = \bm{q}^\top\tanh(\bm{W}\bm{z}_i^r+\bm{b}),
\end{equation}
where $\bm{q}\in \mathbb{R}^{d}$ denotes the attention vector, $\bm{W} \in \mathbb{R}^{d\times d}$ is the weight matrix parameter, and $\bm{b}\in \mathbb{R}^d$ is the bias vector. 
For node \(i\), the weight of each view embedding $\bm{z}_i^r$ can be obtained by:
\begin{equation}
    \beta^r_i=\frac{\exp{(w_i^r)}}{\sum_{r=1}^{|\mathcal{R}|}\exp{(w_i^r)}}.
\end{equation}
Then, the fused representation is obtained by taking weighted average of view representations:
\begin{equation}
    \bm{z}_i= \sum\nolimits_{r=1}^{|\mathcal{R}|}\beta_i^r\bm{z}_i^r.
\end{equation}
These fused representations can be used for downstream tasks.

\section{Experiments}
In this section, we conduct experiments on three real-world datasets to evaluate our proposed method.

\subsection{Experimental Setup}
\noindent \textbf{$\diamondsuit$ Datasets.}
We conduct experiments on ACM\footnote{\url{https://www.acm.org/}}, IMDB\footnote{\url{https://www.imdb.com/}}, and DBLP\footnote{\url{https://aminer.org/AMinerNetwork/}}. ACM contains item nodes with two types of relations: Paper--Author--Paper (P--A--P) and Paper--Subject--Paper (P--S--P).
IMDB contains movie nodes with Movie--Actor--Movie (M--A--M) and Movie-Director-Movie (M--D--M) relations.
DBLP has paper nodes with Paper--Paper--Paper (P--P--P), Paper--Author--Paper (P--A--P), and Paper--Author--Term--Author--Paper (P--A--T--A--P) relations. 
For fair comparison, we follow the same data preprocessing as in DMGI \cite{Park:2020jj} for all datasets, whose statistics are shown in Table \ref{tab:datasets}. 

\begin{table}[t]
	\centering
	\caption{Statistics of datasets used in experiments.}
	\resizebox{\linewidth}{!}{\begin{tabular}{cccccc}
			\toprule
			Dataset & Relations & \#Nodes  & \#Edges & \#Attributes& \#Classes \\
			\midrule
			\multirow{2}[0]{*}{ACM} & P--S--P & \multirow{2}[0]{*}{3,025} &2,210,761 &\multirow{2}[0]{*}{1,830}&\multirow{2}[0]{*}{3} \\
			&P--A--P& & 29,281&& \\
			\midrule
		    \multirow{2}[0]{*}{IMDB} & M--A--M &\multirow{2}[0]{*}{3550} & 66,428&\multirow{2}[0]{*}{1,007}& \multirow{2}[0]{*}{3} \\
		    & M--D--M  &       &    13,788   &  \\
		    \midrule
             \multirow{3}[0]{*}{DBLP} & P--A--P & \multirow{3}[0]{*}{7,907} &144,783 &\multirow{3}[0]{*}{2,000}&\multirow{3}[0]{*}{4} \\
             &P--P--P & & 90,145&& \\
             &P--A--T--A--P & & 57,137,515& \\
			\bottomrule
	\end{tabular}}

	\label{tab:datasets}
\end{table}

\noindent \textbf{$\diamondsuit$ Baselines.}
The baselines include (a) homogeneous network models are {DeepWalk} \cite{perozzi2014deepwalk}, {Node2Vec} \cite{grover2016node2vec}, {ANRL} \cite{zhang2018anrl}, {GCN} \cite{Kipf:2017tc}, {GAT} \cite{velivckovic2018graph}, {DGI} \cite{Velickovic:2019tu}, and {GraphCL} \cite{you2020graph}, (b) heterogeneous network models {Metapath2vec} \cite{dong2017metapath2vec} and {HAN} \cite{Wang:2019gv}, and (c) multiview network models {MNE} \cite{zhang2018scalable}, {GATNE} \cite{cen2019representation}, {DMGI} \cite{Park:2020jj}, and HDMI \cite{jing2021hdmi}.

\noindent \textbf{$\diamondsuit$ Implementation details.}
For all methods, we set the embedding size to 64 and default to the recommended hyperparameters settings.
For \themodel, we use the Adam optimizer \cite{kingma2014adam} with the initial learning rate to 0.001, the weight decay to 1e-5, the temperature $\tau$ to 0.7, and set the dropout of view encoder to 0.6. We implement our \themodel in MindSpore.
Following DMGI \cite{Park:2020jj}, we run the evaluation for 50 times and report the averaged performance.
We use a logistic regression and a k-Means model to perform node classification and node clustering on the learned embeddings, respectively. We use Macro-F1 (MaF1) and Micro-F1 (MiF1) as metrics for node classification, and Normalized Mutual Information (NMI) for node clustering. 

\subsection{Performance Comparison}
\label{sec:performance}

We first report the performance of all compared methods on node classification and node clustering tasks. Table \ref{tab:classification} summaries the results. Our \themodel consistently achieves the best performance on three datasets. Compared with the strong baselines DMGI and HDMI, \themodel obtains the most noticeable performance improvement. This verifies that our framework has strong capabilities to utilize different graph views. \themodel is also competitive with semi-supervised models, i.e., HAN, GAT, and GCN, which shows the superiority of our framework in the training of view encoder and  multiview aggregator. Traditional baselines MNE and Metapath2vec are inferior to that of attribute-aware network methods, such as HAN, ANRL, DMGI, and HDMI, on most datasets. This indicates that the node attributes are necessary for multiview network embedding. Furthermore, most multiview methods, such as HDMI, DMGI, GATNE, and MNE, generally outperform single-view methods. This verifies the necessity of modeling multiple relations. GraphCL and DGI, as the contrastive learning methods, perform the best among single-view network embedding methods in most datasets. This result demonstrates the superiority of mutual information optimization over proximity reconstruction in representation learning.

\begin{table}[t]
	\centering
	\caption{Performance comparison of different models. The highest and second-to-best performance is highlighted in boldface and underlined respectively.}
	\resizebox{\linewidth}{!}{
		\begin{tabular}{ccccccccccccc}
			\toprule
			\multirow{2.5}[0]{*}{Method}  & \multicolumn{3}{c}{ACM} & \multicolumn{3}{c}{IMDB} & \multicolumn{2}{c}{DBLP} \\
			\cmidrule(lr){2-4} \cmidrule(lr){5-7} \cmidrule(lr){8-10}  
			& MaF1 &MiF1 &NMI& MaF1 &MiF1&NMI& MaF1 &MiF1&NMI \\
			\midrule
			Deepwalk & 0.739 & 0.748& 0.310 & 0.532&0.55 &0.117& 0.533& 0.537& 0.348\\
			node2vec & 0.741 &0.749& 0.309& 0.533 &0.55 &0.123&0.543 &0.547& 0.382\\
			ANRL     & 0.819 &0.820&0.515& 0.573 &0.576 &0.163& 0.770 &0.699& 0.332 \\
			GCN/GAT  & 0.869 &0.870& 0.671& 0.603 &0.611&0.176& 0.734 &0.717& 0.465 \\
			DGI      & 0.881 &0.881&0.640& 0.598 &0.606 &0.182& 0.723&0.720& 0.551\\
			GraphCL   &0.892&0.894&0.656& 0.613&0.624 & 0.183 & 0.736 &0.722&0.562\\
			\midrule
			Metapath2vec   & 0.752 &0.758&0.314&0.546&0.574 &0.144 & 0.653&0.649&0.382\\
			HAN    & 0.878 &0.879& 0.658& 0.599 &0.607&0.164 & 0.716&0.708 &0.472 \\
			\midrule
			MNE   & 0.792&0.797 &  0.545& 0.552 &0574&0.013& 0.566 &0.562&0.136 \\
			GATNE & 0.846& 0.841& 0.521&0.494&0.504&0.048& 0.673 &0.665&0.436 \\
			
			DMGI  & 0.898&\underline{0.898}  &{0.687}& \underline{0.648}&\underline{0.648} &{0.196}& 0.771 &0.766&0.409 \\
			DMGI-attn & {0.887} &{0.887}& \underline{0.702}& 0.602&0.606&0.185 &{0.778} &\underline{0.770}& \underline{0.554} \\
			HDMI & \underline{0.895} &{0.894} &0.657 & 0.601&0.610&\underline{0.197}&\underline{0.805}&\underline{0.795}&0.544\\
			\midrule
			\textbf{\themodel} & \textbf{0.907} & \textbf{0.906}& \textbf{0.726}&\textbf{0.672}& \textbf{0.675} &\textbf{0.211} &\textbf{0.812}& \textbf{0.798} &  \textbf{0.623}\\
			\bottomrule
		\end{tabular}
	}
	\label{tab:classification}
\end{table}

\subsection{Ablation Studies}

To investigate the effects of the contrastive objectives, view encoder, and multiview aggregator, we compare \themodel with five variants.
\emph{CRE\textsubscript{V}-mean} and \emph{CRE\textsubscript{V}-max} set the operator as mean and max in the view encoder, respectively.
\emph{CRE\textsubscript{M}-mean} and \emph{CRE\textsubscript{M}-max} set the operator as mean and max in the multiview aggregator, respectively.
\emph{CRE\textsubscript{C}-ori} excludes the inter-view InfoMin objective.
The results are shown in Table \ref{ablation study}.
From the table, we see that {CRE$_V$-mean} and {CRE$_V$-max} perform worse than {CRE$_M$-mean} and {CRE$_M$-max} on most datasets, especially for node clustering, which suggests that the view encoder plays a more important role compared to the multiview aggregator.
The performance of {CRE$_M$-mean} and {CRE$_M$-max} is not significantly different from that of \themodel in ACM and IMDB. However, the performance of {CRE$_M$-max} is worse in DBLP. The reason is that DBLP data is more complicated than ACM and IMDB, as shown by the fact that DBLP have more relations. The max aggregator tends to ignore multiplicities than the attention and mean aggregator \cite{xu2018powerful}. 
\themodel outperforms {CRE$_C$-ori} in most cases, which demonstrates that our inter-view InfoMin could supplement the view fusion InfoMax objective.

\begin{table}[t]
	\centering
	\caption{Performance of different model variants.}
	\resizebox{\linewidth}{!}{
	\begin{tabular}{cccccccccccc}
	\toprule
	\multirow{2.5}[0]{*}{Variant} & \multicolumn{3}{c}{ACM}& \multicolumn{3}{c}{IMDB} & \multicolumn{3}{c}{DBLP}  \\
	\cmidrule(lr){2-4} \cmidrule(lr){5-7} \cmidrule(lr){8-10} 
	 & MaF1 &MiF1 &NMI& MaF1&MiF1 &NMI & MaF1&MiF1 &NMI \\
	\midrule
    {CRE$_V$-mean} & 0.786 &0.778 & 0.394& 0.519 &0.546 & 0.056 &0.801&0.795&0.516\\
	{CRE$_V$-max} & 0.824 &0.828 &0.529 & 0.551 &0.562 &0.015 &0.810&0.796&0.516  \\

	{CRE$_M$-mean}  & 0.896 &0.896& 0.714& 0.672 &0.673 & 0.196 &0.803&0.783&0.623 \\
	{CRE$_M$-max} & 0.905 &0.899 & 0.723 & 0.671 &0.674 &0.203 &0.792&0.780&0.606\\
	
    {CRE$_C$-ori}& 0.894 &0.893&0.725& 0.657 &0.661 &\textbf{0.216} &0.795&0.775&0.519 \\
    \midrule
	\textbf{\themodel} &\textbf{0.907}&\textbf{0.906}&\textbf{0.726} &\textbf{0.672}&\textbf{0.675}&0.211 &\textbf{0.812}&\textbf{0.798} &\textbf{0.623}\\
	\bottomrule
	\end{tabular}
	}
	\label{ablation study}
\end{table}

\subsection{Visualization}

\begin{figure}[t]
	\centering
	\subfloat[HDMI]{\includegraphics[width=0.42\linewidth]{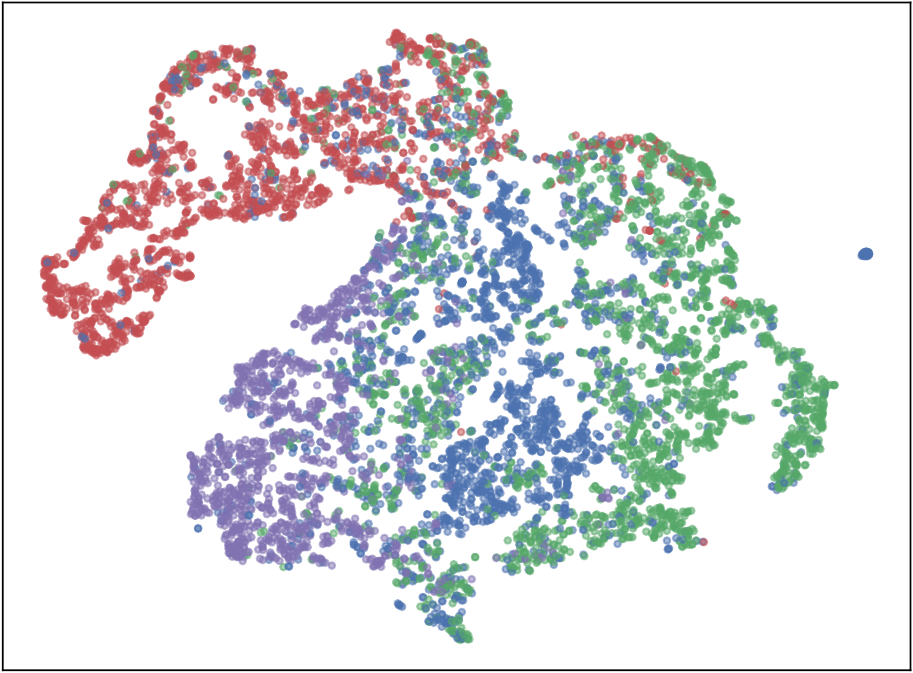}
		\label{fig_first_case}}\qquad
	\subfloat[CREME]{\includegraphics[width=0.42\linewidth]{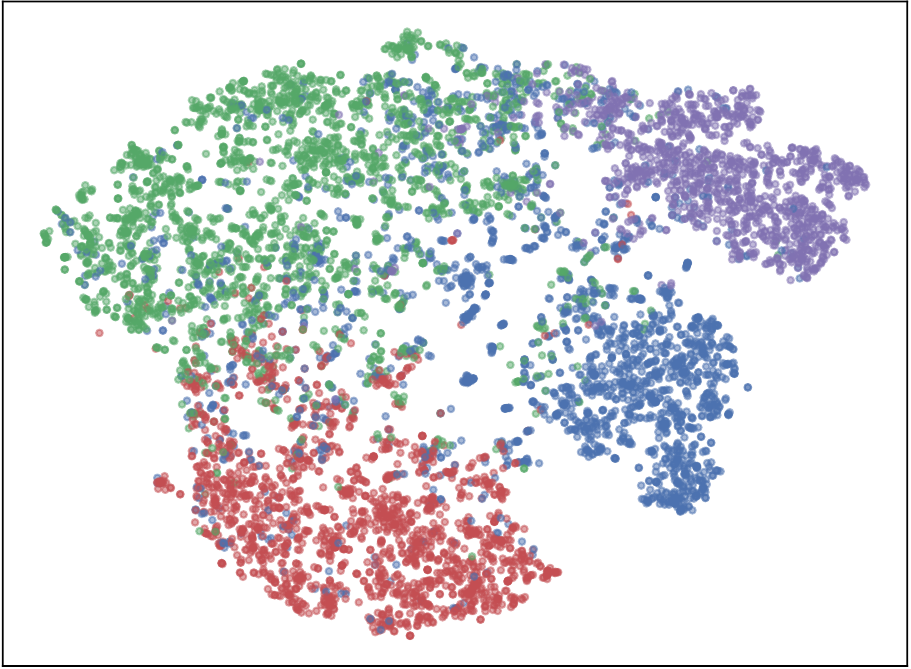}
		\label{fig_second_case}}
	\caption{Visualization of the learned node embedding by HDMI and \themodel on DBLP.}
	\label{sne}
\end{figure}
To provide a qualitative evaluation, we map the node embedding of the DBLP network learned by \themodel and HDMI into a 2D space using the t-SNE algorithm \cite{van2008visualizing} and plot them in Figure \ref{sne}.
We find that \themodel exhibit more distinct boundaries and clusters than HDMI.
Moreover, the Silhouette scores \cite{rousseeuw1987silhouettes} of the embeddings obtained by HDMI and \themodel are 0.11 and 0.28 (the higher, the better), respectively, which once again verifies that \themodel can learn informative node embeddings.

\section{Conclusion}
In this work, we have proposed a novel contrastive learning framework for unsupervised learning of multiview networks. 
In our framework, we propose two contrastive objectives through optimizing mutual information between different views, fusion view InfoMax and inter-view InfoMin, which distills information from every relation view and promotes discriminative view representations.
Extensive experiments on three real-world multiview networks verify the effectiveness of \themodel.

\begin{acks}
This work is supported by National Natural Science Foundation of China (62141608 and U19B2038) and CAAI--Huawei MindSpore Open Fund.
\end{acks}

\bibliographystyle{ACM-Reference-Format}
\balance
\bibliography{cikm2022.bib}

\end{document}